\documentclass[letterpaper, 10 pt, conference]{ieeeconf}  

\IEEEoverridecommandlockouts                              

\usepackage{amsmath} 
\usepackage{amssymb}  
\usepackage{booktabs}
\usepackage{multirow}
\usepackage{makecell}

\usepackage[caption=false,font=normalsize,labelfont=sf,textfont=sf]{subfig}
\usepackage{caption}
\usepackage[capitalise]{cleveref}
\usepackage{graphicx}

\usepackage{xcolor}

\usepackage{calc}
\usepackage{fp}

\definecolor{mygreen}{RGB}{68, 152, 92}
\definecolor{mypink}{RGB}{217, 217, 255} 

\newcommand{\percentage}[2]{%
    \FPeval{\result}{round(#1/#2*100,1)}%
    \result\%
}
\newcommand{\improvement}[1]{\textcolor{mygreen}{\scriptsize +#1\%}}

\usepackage{todonotes}

\title{\LARGE \bf
MobRT: A Digital Twin-Based Framework for Scalable Learning in Mobile Manipulation
}

\author{Yilin Mei, Peng Qiu, Wei Zhang, WenChao Zhang, Wenjie Song$^{\dagger}$
\thanks{This work was partly supported by Program for National Natural Science Foundation  of  China  (Grant  No.  62373052),  Beijing  Natural  Science Foundation (Grant No. 4252051), and in part by the National Key Laboratory of Science and Technology on Space Born Intelligent Information Processing TJ-01-22-09.}
\thanks{$^{*}$Wenjie Song is with School of Automation, Beijing Institute of Technology, Beijing 100081, China,
        {\tt\small songwi@bit.edu.cn}}%
}

\begin{document}

\maketitle
\thispagestyle{empty}
\pagestyle{empty}

\begin{abstract}

Recent advances in robotics have been largely driven by imitation learning, which depends critically on large-scale, high-quality demonstration data. However, collecting such data remains a significant challenge—particularly for mobile manipulators, which must coordinate base locomotion and arm manipulation in high-dimensional, dynamic, and partially observable environments. Consequently, most existing research remains focused on simpler tabletop scenarios, leaving mobile manipulation relatively underexplored. To bridge this gap, we present \textit{MobRT}, a digital twin-based framework designed to simulate two primary categories of complex, whole-body tasks: interaction with articulated objects (e.g., opening doors and drawers) and mobile-base pick-and-place operations. \textit{MobRT} autonomously generates diverse and realistic demonstrations through the integration of virtual kinematic control and whole-body motion planning, enabling coherent and physically consistent execution. We evaluate the quality of \textit{MobRT}-generated data across multiple baseline algorithms, establishing a comprehensive benchmark and demonstrating a strong correlation between task success and the number of generated trajectories. Experiments integrating both simulated and real-world demonstrations confirm that our approach markedly improves policy generalization and performance, achieving robust results in both simulated and real-world environments.

\end{abstract}

\section{Introduction}

The deployment of robotic systems in unstructured environments, including homes, hospitals, and warehouses, requires capabilities that extend beyond fine-grained manipulation. To accomplish everyday activities, for example opening cabinets, loading dishwashers, or organizing clutter, robots must integrate reliable object interaction with robust navigation in cluttered, constrained, and dynamic spaces. Mobile manipulators, by coordinating whole-body movements that involve the mobile base, arm, and sometimes even the torso, are particularly well-suited to these challenges and hold great potential for providing practical robotic assistance in daily life.

\begin{figure}[!htbp]
    \centering
    \includegraphics[width=1.0\linewidth]{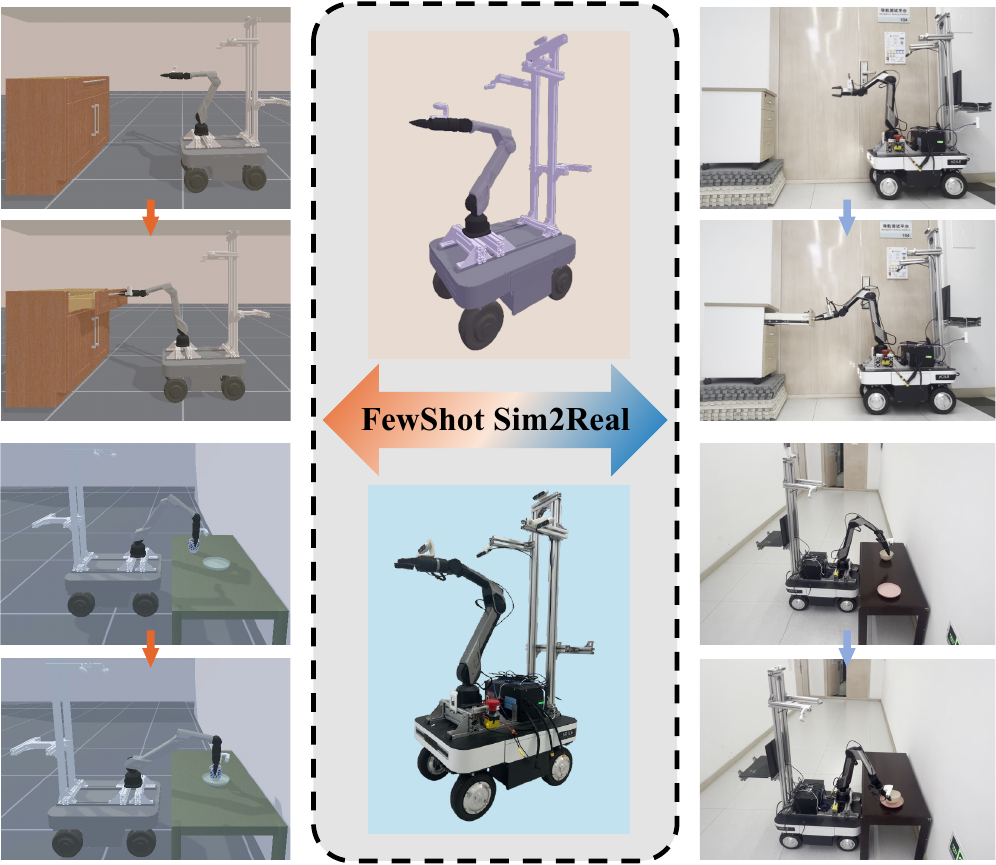}
    \caption{\textbf{FewShot Sim2Real with MobRT.} By leveraging 300 simulated demonstrations and only 20 real ones, mobile manipulators learn articulated-object interaction and mobile-base pick-and-place tasks with successful sim-to-real transfer.}
    \vspace{-0.6cm}
    \label{fig:abstract_overall}
\end{figure}

In recent years, learning-based approaches, particularly imitation learning, have significantly advanced the capabilities of robotic systems, making the deployment of mobile manipulators in these unstructured environments increasingly feasible. By learning from human demonstrations, robots can acquire complex behaviors that would be difficult to hand-engineer. However, these approaches require large-scale, high-quality demonstration datasets for effective generalization. Collecting such data for mobile manipulation tasks remains especially challenging. Traditional methods typically rely on expert teleoperation, which is time-consuming, labor-intensive, and dependent on robot-specific hardware \cite{aloha, mobile_aloha, gello, bidex, behavior_robot_suite}. To reduce these burdens, recent research has investigated remote teleoperation via virtual reality devices \cite{qin2023anyteleop, bunny, dass2024telemoma, zero_cost_teteop}. Although such approaches alleviate specific practical constraints, their continued reliance on manual demonstrations imposes inherent limitations on scalability and generalization. To further reduce the reliance on human effort, simulation-based data generation has become increasingly popular. In simulated environments, demonstration trajectories can be generated using scripted policies, motion planning, or other automated methods, enabling large-scale data collection with minimal human intervention. Benefiting from advancements in simulation techniques such as ray tracing and improved physics engines, the sim-to-real transfer process has become more feasible, although discrepancies between simulated and real-world dynamics still present challenges. However, most existing simulation frameworks primarily focus on fixed-base manipulators and tabletop tasks \cite{mandlekar2023mimicgen, jiang2024dexmimicgen, mu2025robotwin}, offering limited support for mobile manipulation scenarios that require tight coordination between locomotion and manipulation.

To address the aforementioned challenges, we propose \textbf{MobRT}, a scalable data generation and learning platform for mobile manipulators. MobRT offers a simulation environment that enables comprehensive evaluation of whole-body mobile manipulation, supporting scenarios such as articulated-object interaction (e.g., doors and drawers) and coordinated pick-and-place tasks involving the mobile base. By incorporating virtual kinematic control and whole-body motion planning, MobRT can autonomously generate large-scale, diverse, and realistic demonstrations with smooth and physically consistent execution, facilitating the training of manipulation policies with improved generalization. To systematically evaluate the effectiveness of our generated dataset, we establish a comprehensive benchmark and evaluate multiple baseline algorithms. Through extensive experiments conducted in both simulation and real robotic platforms, we validate the utility and generalization of our data. In addition, we propose an additional baseline control policy based on a Transformer neural network trained within the Diffusion Policy framework. Our experiments show that this policy outperforms other tested baselines and is better suited as a baseline model specifically for mobile manipulators.

In summary, our key contributions are as follows.
\begin{enumerate}
    \item \textbf{MobRT Framework}: We develop MobRT, a flexible framework for scalable data generation and learning tailored to mobile manipulators, enabling high-fidelity simulation of complex whole-body mobile manipulation scenarios.
    \item \textbf{Benchmark Evaluation}: We create a comprehensive benchmark based on data generated by MobRT, and use it to evaluate multiple baseline methods through extensive experiments in both simulated and real-world settings.
    \item \textbf{Baseline Policy}: We propose a new baseline control policy that outperforms existing baselines and provides a stronger foundation specifically for mobile manipulation tasks.
\end{enumerate}

\section{Related Works}

\subsection{Mobile Manipulation}
Despite advances in fixed-base arms, mobile manipulation remains underexplored due to its high dimensionality arising from the need to plan motions for both the base and arm simultaneously. Previous works often separate these tasks by first moving the base to a predefined position before initiating manipulation \cite{xiong2024adaptive, zhi2024closed}. Such approaches limit flexibility and fail to use the base's full capabilities. Using such motion planning methods also results in unnatural and incoherent simulation data; for example, when opening doors or drawers, the base should move in coordination with the arm rather than the arm acting alone. MobRT solves this by using whole-body control to generate smoother, more coordinated motions for articulated object manipulation.

\subsection{Demonstration Generation}
High-quality demonstrations for robotic tasks are typically collected through human teleoperation. However, this approach is both costly and time-consuming, especially for mobile manipulators, since operators must control both the robotic arm and the mobile base simultaneously, substantially complicating the control process. Efforts to simplify teleoperation have explored whole-body control methods, for instance by using only the end-effector’s trajectory to guide the manipulator’s motion \cite{chi2024umi, ha2024umionlegs, zero_cost_teteop}. Nevertheless, acquiring large-scale and diverse demonstration data still necessitates considerable operator involvement across many scenarios. Therefore, simulation based data generation has emerged as a promising alternative, offering a cost-effective and scalable means to produce rich datasets. Among existing works, the two most closely related to our approach are MoMaGen \cite{li2025momagen} and RoboTwin \cite{mu2025robotwin}. MoMaGen makes a notable contribution by extending the X-Gen paradigm to generate data tailored for mobile manipulators and by producing diverse trajectories through replaying human demonstrations. While MoMaGen makes progress in adapting trajectory replay to mobile manipulators, its omission of whole-body control restricts its applicability to complex tasks. In addition, its dependence on replayed demonstrations may introduce biases that impede the discovery of optimal trajectories. In contrast, RoboTwin has advanced dual-arm coordination and provides a high-quality framework for studying fine-grained desktop manipulation. However, its scope remains restricted to tabletop settings and does not address the distinct challenges inherent to mobile manipulation.

\subsection{Robot Manipulation Learning}
Although many methods have explored autonomous skill acquisition for robots via reinforcement learning in simulation, learning manipulation skills from high-quality human demonstrations remains the most effective approach for mobile manipulators, particularly when dealing with complex tasks. Recent advances in generative model-based imitation learning have enabled robots to perform increasingly complex behaviors \cite{aloha, chi2023diffusionpolicy, jia2024lift3d, ze2024dp3, ze2024idp3, black2410pi0}. Building upon the strengths of these recent imitation learning frameworks, we propose a baseline algorithm specifically tailored for mobile manipulators. Our method integrates the key advantages of existing approaches to enhance learning efficiency and adaptability in tackling complex tasks.

\begin{figure*}[!htbp]
    \centering
    \includegraphics[width=1.0\linewidth]{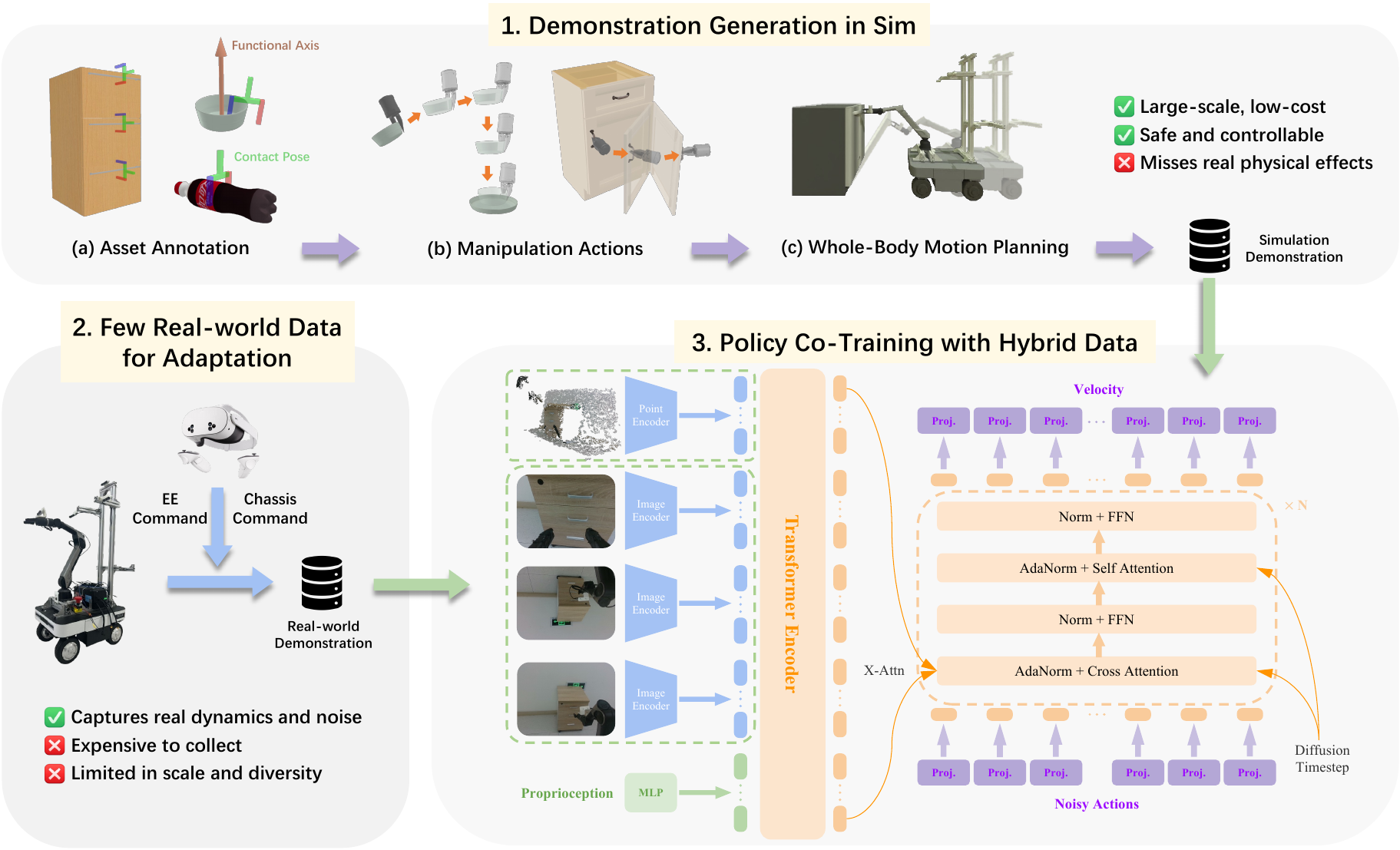}
    \caption{\textbf{Overview of the MobRT pipeline.} The framework integrates three key components: (1) \emph{Demonstration generation in simulation}, where asset annotation, manipulation actions generation, and whole-body motion planning enable large-scale, low-cost data collection; (2) \emph{Adaptation with real-world demonstrations}, where a small set of trajectories captures real dynamics and sensor noise to complement the simulated data; and (3) \emph{Policy training with hybrid data}, where multi-modal encoders process visual and proprioceptive inputs and a transformer-based diffusion policy learns to coordinate mobile-base and arm motions for whole-body tasks.}
    \label{fig:method_overall}
    \vspace{-0.6cm}
\end{figure*}

\section{Method}

In this section, we first describe our approach to data collection in simulation, focusing on two key challenges: generating actions for manipulating specific objects and producing whole-body motion trajectories for the mobile manipulator. We then present our method for distilling the collected trajectories into a policy network. This policy network, proposed as part of our work, serves as a new baseline for mobile manipulator manipulation tasks. The overall workflow of our approach is illustrated in \cref{fig:method_overall}.

\subsection{Demonstration Generation in Simulation}

\textbf{Digital Twin Assets and Annotation: } Creating precise and realistic digital twins of physical objects is a fundamental prerequisite for synthesizing high-fidelity expert data through simulation. To this end, we leverage the PartNet-Mobility \cite{xiang2020sapien} and UniDoorManip \cite{li2024unidoormanip} datasets, which encompass hundreds of articulated objects across various categories, alongside the RoboTwin-OD \cite{chen2025robotwin} dataset, a rich set of rigid-body objects generated via Artificial Intelligence Generated Content (AIGC) from simple 2D RGB images. For the objects in our experimental environment, we adopt the same AIGC-based approach as RoboTwin-OD to generate their digital twin assets. To ensure physically consistent interaction, these assets are further processed using CoACD \cite{wei2022approximate} for convex decomposition. To generate demonstrations, we annotate each object with a 6-DoF grasp pose, and for certain rigid-body objects, such as containers and plates, we additionally define a functional axis that indicates their intended use, such as placement. By combining this annotation with Virtual Kinematic Chains (VKC) \cite{jiao2021consolidating, jiao2021efficient} and predefined motion primitives, we are able to synthesize a diverse set of demonstration trajectories that reflect realistic and varied manipulation behaviors.

\textbf{Generating Manipulating Actions: } One of the key challenges in demonstration generation is planning trajectories for managing interactions between two rigid-body objects or manipulating articulated components, such as placing a container onto a plate, or rotating a lever-style door handle while pulling or pushing the door. Many previous approaches address this challenge by using hard-coded motions, such as manually specifying approximate end-effector movements to open a drawer. These methods are inconvenient and require extensive tuning, which limits their scalability and adaptability. 

\begin{figure}[!htbp]
    \centering
    \includegraphics[width=1.0\linewidth]{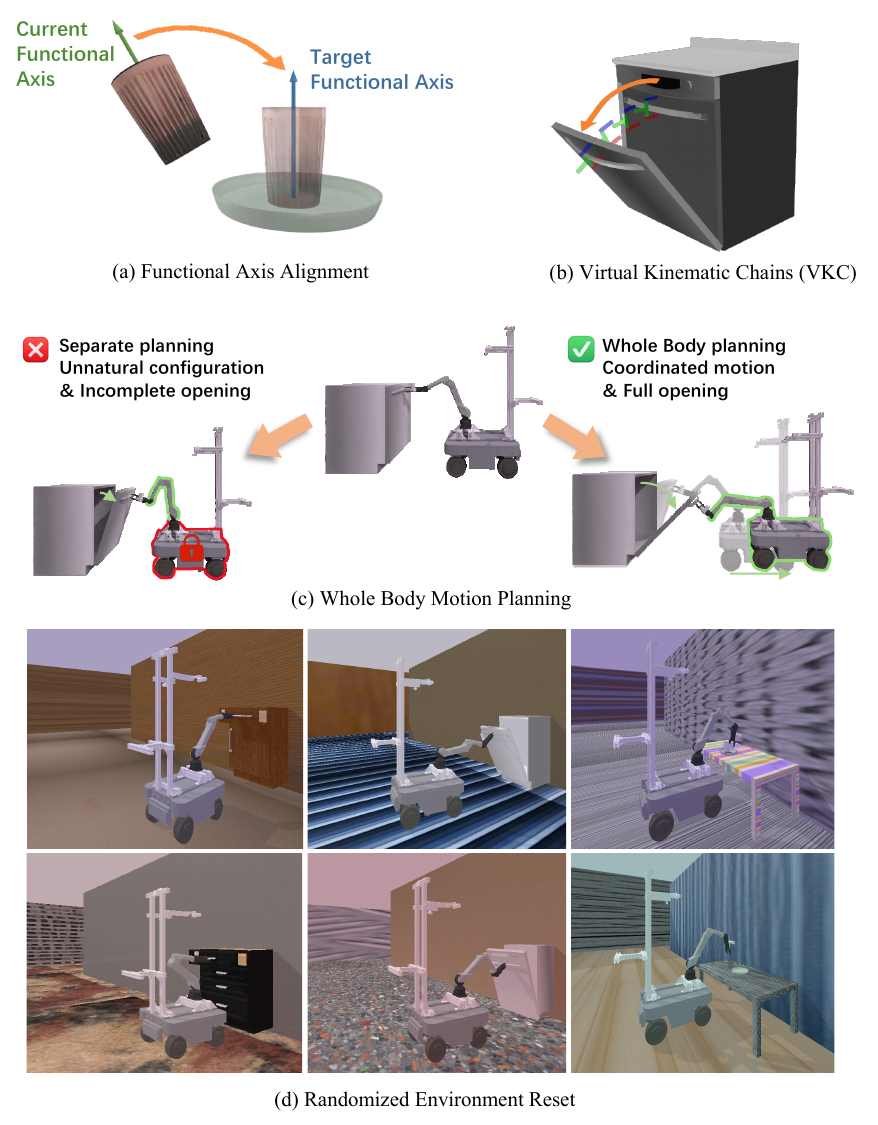}
    \caption{\textbf{Generating simulation demonstrations.} (a). Functional-axis alignment to synthesize pick-and-place actions; (b). VKC to generate articulated-object manipulation; (c). Whole-body Planning vs. Separate Planning for coordinated base–arm motion; (d). Randomized environment reset for diverse, validated data.}
    \vspace{-0.6cm}
    \label{fig:method_detail}
\end{figure}

To address these limitations, we adopt different strategies for each type of interaction. For interactions between rigid-body objects, we simplify the problem by leveraging the annotated functional axis to align the axes of the two objects, as illustrated in \cref{fig:method_detail}.(a). For tasks involving articulated components, we employ the Virtual Kinematic Chains (VKC) framework to generate the corresponding end-effector pose trajectories for manipulation, as shown in \cref{fig:method_detail}.(b). 

In the VKC framework, the manipulation process proceeds as follows: once the robot’s end-effector reaches the target pose, the gripper closes to establish a rigid connection with the articulated link. The simulator then provides access to the articulation’s kinematic model. For a given articulated link, we first identify its associated joint and, depending on the task, sample a sequence of intermediate joint values from the current state to the desired target. Using the articulation’s kinematic model, we compute the corresponding pose trajectory of the articulated link \(\{\mathcal{T}_{\text{link}}\}_{t=1}^T\). This trajectory is subsequently mapped to an end-effector pose trajectory \(\{\mathcal{T}_{\text{eef}}\}_{t=1}^T = \{\mathcal{T}_{\text{link}}(\theta) * \mathcal{T}_{\text{link}}^{-1}(\theta_{\text{init}}) * \mathcal{T}_{\text{eef}}^{\text{init}}\}_{t=1}^T\), thereby reducing the manipulation of articulated components to an end-effector trajectory-tracking problem.

In this way, the two most common types of manipulation tasks, namely pick-and-place operations and interactions with articulated objects, can be represented as sequences of key end-effector poses, which can then be executed through motion planning.

\textbf{Whole-Body Motion Planning}: Producing coherent and physically consistent manipulation behaviors in mobile manipulators requires integrated whole-body motion planning, which enables the coordinated control of both the mobile base and the robotic arm, as illustrated in \cref{fig:method_detail}.(c). To achieve this, we compute a whole-body trajectory \( x[1, T] \) that moves the end-effector from its current pose to the target pose via an optimization process defined as: 

\[
\begin{aligned}
\min_{x[1, T]} \quad & \mathcal{C}_{\text{eef}}(x_T, \mathcal{T}_{\text{eef}}) + \sum_{t=1}^T \left[\mathcal{C}_{\text{smooth}} (x(t)) + \mathcal{C}_{\text{base}}(x(t)) \right] \\
\text{s.t.} \quad & x(1) = x_{\text{init}} \\
& x_{\text{min}} \leq x(t) \leq x_{\text{max}} \quad \forall t \\
\end{aligned}
\]

The formulation aims to drive the end-effector pose to the goal pose \(\mathcal{T}_{\text{eef}}\) while minimizing control effort \( \mathcal{C}_{\text{smooth}} \) and maximizing compliance to chassis soft constraints \( \mathcal{C}_{\text{base}} \). These chassis soft constraints include maintaining a fixed orientation of the chassis throughout the whole-body motion, such as when pushing a door through a frame, and minimizing collision cost, for example, by avoiding collision with a table during pick-and-place operations. To further improve execution efficiency, the computed whole-body trajectories are post-processed using the Time-Optimal Path Parameterization (TOPP-RA) algorithm \cite{pham2018new}. This approach generates time-optimal trajectories along the planned path while adhering to joint velocity and acceleration limits, ensuring that the resulting motions are dynamically feasible and exhibit enhanced continuity.

Furthermore, in addition to whole-body control, we also implement separate trajectory generation for the chassis and the arm, enabling more flexible and diverse motion types. Together with gripper open and close actions, these motions form a set of motion primitives. Using these motion primitives, diverse task workflows can be effectively composed.

\textbf{Environment Reset: } Some trajectories may fail to complete the task for various reasons, such as motion planning failure, collisions with the environment during execution, or the failure of the end-effector to establish a stable grasp on the object. Therefore, before resetting the environment and saving the data, we validate each trajectory to determine whether it successfully accomplishes the task. For pick-and-place tasks, we check whether the distance between the two objects falls below a predefined threshold; for articulated-object manipulation tasks, we verify whether the target joint reaches the specified angle. This simple filtering step effectively eliminates a large number of invalid trajectories. At each environment reset, the base of the mobile manipulator is placed at the center of the world frame. The reset process involves extensive randomization, including the object's position, orientation, and size; the initial joint configuration of the robotic arm; the textures of the ground and selected objects; and the lighting conditions. Textures from \cite{chen2025robotwin} and \cite{teoh2024green} are used to increase visual variation. These rich randomization strategies ensure diversity in the collected data and improve the robustness of subsequent learning.

\subsection{Policy Learning}

After collecting the demonstration data, the key challenge is how to distill it into a usable policy. Building on insights from prior work, we design a novel baseline policy that leverages a Transformer-based neural network trained under the Diffusion Policy framework, which offers an expressive formulation for imitation learning and supports multimodal inputs, facilitating easy extension of the model.

\textbf{Flow Matching for Action Generation: } In our work, we adopt \textbf{Flow Matching}, a variant of diffusion models based on optimal transport theory, to train our policy. Unlike traditional diffusion models that rely on simulating a full stochastic process, Flow Matching learns a time-dependent vector field that transports samples from noise to expert actions in fewer steps, enabling faster and more stable inference. 

Formally, let \( \{\tau_i\}_{i=1}^N \) represent our dataset of demonstrations, where each \( \tau_i = \{(o_1^i, a_1^i), \dots, (o_T^i, a_T^i)\} \) is a sequence of \( T \) observations and their corresponding actions. Here, \( o_t^i \) represents an observation at time \( t \), and \( a_t^i \) is the corresponding action. Our goal is to optimize a policy \(\pi_\theta\) such that the predicted actions \(\pi_\theta(o_t^i)\) closely match the expert actions \(a_t^i\) across the entire dataset. To improve temporal consistency, we apply Action Chunking as proposed in \cite{aloha}. 

Given an action chunk \(\mathbf{A} = [a_t, a_{t+1}, \dots, a_{t+H-1}]\) and a Gaussian noise sample \(\mathbf{Z} = [z_t, z_{t+1}, \cdots, z_{t+H-1}]\), we construct interpolated samples as follows:

\[
\mathbf{X}_\tau = \tau \mathbf{A} + (1 - \tau) \mathbf{Z}
\]

where \(\tau \in [0, 1]\) represents the flowing matching timestep, controlling the interpolation between pure noise and expert actions.

We then train a policy \( \pi_\theta \) with output \(v(\tau, \mathbf{X}_t, o_t)\) to match the optimal flow by minimizing the following loss:

\[
\mathcal{L}_{FM} = \mathbb{E}_{\tau, \mathbf{Z}, \mathbf{A}} \lVert v(\tau, \mathbf{X}_t, o_t) - (\mathbf{A} - \mathbf{Z}) \rVert ^ 2
\]

This formulation allows us to train with 100 diffusion steps, but perform inference using only 10 steps, without sacrificing performance.

\textbf{Architecture details: } Our policy model consists of a multi-modal encoder and a Transformer-based decoder. RGB observations are processed using a pre-trained ResNet-18 backbone. For point cloud inputs, outputs from different cameras are first transformed into a unified coordinate frame based on their relative poses, after which all point clouds are fused and jointly encoded using a shared PointEncoder, following the approach of \cite{jia2024lift3d}. The proprioceptive state $s_t$ and the noisy action sequence $\mathbf{A}$ are each encoded through separate MLPs. Unlike prior works that employ a full Transformer decoder block with self-attention, cross-attention, and feed-forward layers, we follow \cite{shukor2025smolvla} and alternate between self-attention and cross-attention: cross-attention enables interaction with multi-modal encoder outputs, while self-attention smooths the action sequence. The flow time $\tau$ is mapped into timestep embeddings and integrated via AdaNorm. The overall model architecture is illustrated in \cref{fig:method_overall}.

\textbf{Co-training with Real-world Data: } While our framework can generate a large amount of demonstration data in simulation, these trajectories are primarily produced through motion planning. As a result, they fail to fully capture the complexities of real-world robot systems, such as dynamic behaviors, sensor noise, and actuation delays. This discrepancy often limits the effectiveness of direct sim-to-real transfer. Specifically, actions that can be executed perfectly in simulation may exhibit steady-state errors or control inaccuracies in real-world environments. 

To bridge this gap, we supplement the synthetic data with real-world demonstrations. Let \( D_{real}^m \) denote the aggregated real-world demonstrations for a task \( m \), and \( D_{sim}^m \) denote the simulations. The training objective for the policy \( \pi_\theta^m \) for a task \( m \) is formulated as:

\[
\mathbb{E}_{(o^i, a^i) \sim D_{sim}^m } [\mathcal{L}(a^i, \pi_\theta^m(o^i)] + \mathbb{E}_{(o^i, a^i) \sim D_{real}^m} [\mathcal{L}(a^i, \pi_\theta^m(o^i)]
\]

where $\mathcal{L}$ is the Flow Matching loss function as defined earlier. During training, we sample from both the real-world demonstrations \( D_{real}^m \) and the simulations demonstrations \( D_{sim}^m \) with equal probability.

\textbf{Training details: } We train our policy model using an RTX A6000 GPU with a batch size of 64 for 100K steps. We employ the Adam optimizer with a weight decay of 1e-6. The learning rate schedule consists of a linear warmup over the first 500 steps, reaching a peak learning rate of 1e-4, followed by a cosine decay to a minimum learning rate of 1e-6. Gradient clipping is applied with a maximum norm of 10.

\section{Experimental Results}

We design experiments to evaluate the effectiveness of MobRT from two key perspectives: (1) To validate the quality of the data generated by MobRT, we construct a benchmark that demonstrates its applicability in evaluating policy generalization in mobile robot tasks; (2) To show that our proposed policy significantly outperforms baseline algorithms across a range of mobile manipulation tasks. In the benchmark experiments, we investigate the correlation between the number of MobRT-generated trajectories and the success rates of baseline algorithms. In the real-world experiments, we assess how training with a combination of simulated and real-world data impacts model performance. These experiments are designed to rigorously evaluate both the effectiveness of MobRT-generated data and the performance of our proposed policy.

\subsection{Experiment Setups}
\subsubsection{Simulation}

We employ ManiSkill3 as the physics simulator due to its GPU-based visual data collection capabilities, enabling efficient acquisition of RGB-D and segmentation data. With ray tracing, it produces highly realistic visual effects, making it well-suited for building a mobile robot data collection system.

Articulated object manipulation tasks are constructed using PartNet-Mobility and UniDoorManip, with objects selected from two categories: cabinets, and dishwashers. For pick-and-place tasks, we adopt the RoboTwin-OD dataset and design similar tasks to \cite{mu2025robotwin}, but with the added requirement of base movement during manipulation.


\begin{figure}[!htbp]
    \centering
    \includegraphics[width=0.5\textwidth]{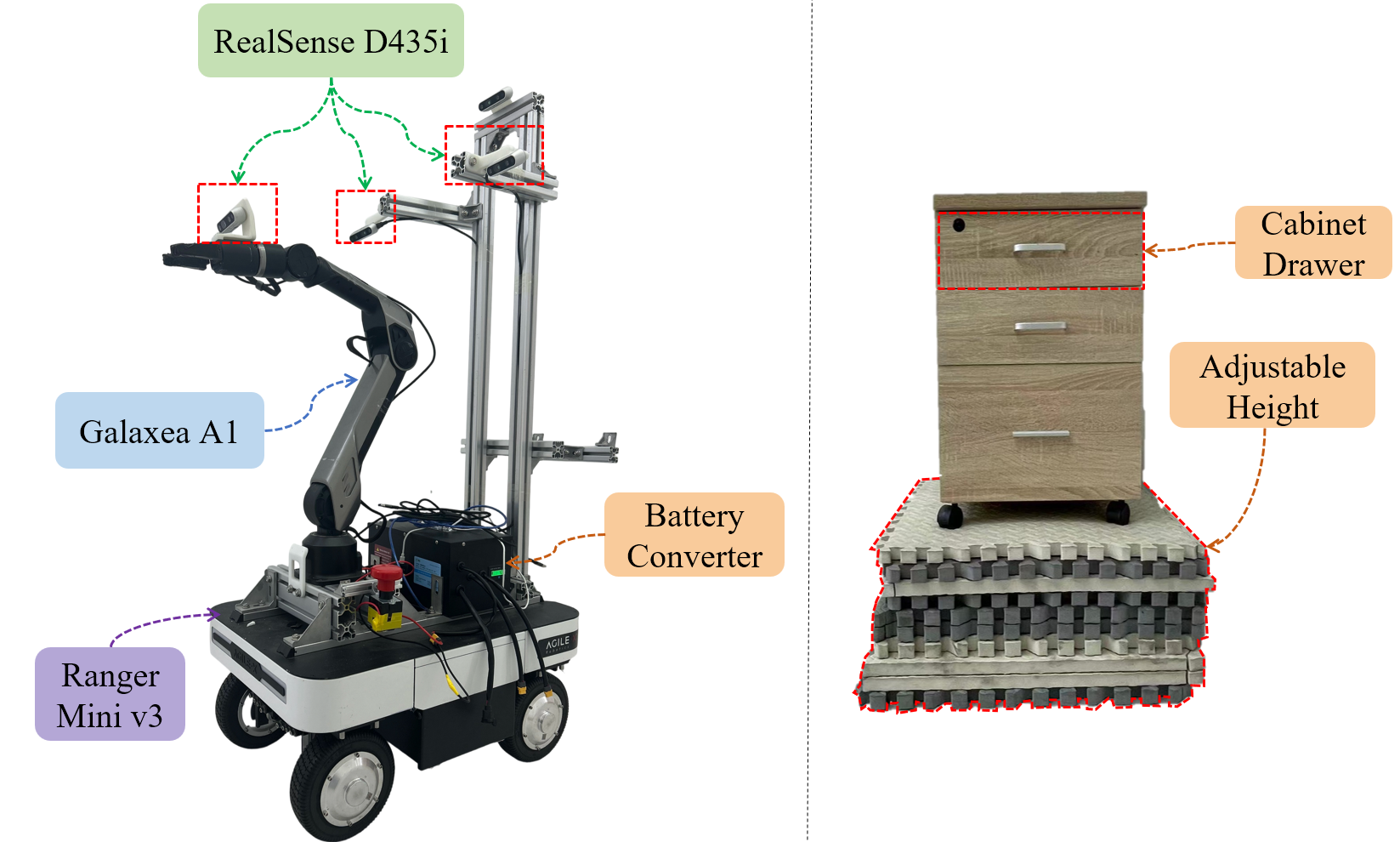}
    \caption{\textbf{Real-World experimental environment and Illustration of our robot platform.} To test the robustness of the system, the drawers are placed at variable heights.}
    \vspace{-0.6cm}
    \label{fig:robot_system}
\end{figure}

\subsubsection{Real-World}

In addition to standard domain randomization, we follow \cite{dalal2024local} to perturb the camera's depth output for greater realism. Specifically, we simulate edge artifacts near depth discontinuities and random holes in the depth map, mimicking noise from real sensors such as Intel RealSense. To mitigate these effects, we apply voxel downsampling and statistical outlier removal to the resulting point clouds, consistently for both simulated and real data, thereby aligning their distributions and enhancing sim-to-real transfer.

As illustrated in Fig.~\ref{fig:robot_system}, our real-world mobile manipulator platform consists of a Galaxea A1 robotic arm mounted on a Ranger Mini 3 mobile base. Three Intel RealSense D435i cameras are deployed: two are positioned above the rear of the robot on the left and right sides to capture global environmental information, while the third is mounted on the robot's wrist to provide close-range observations.

The robot is controlled via ROS using the Galaxea A1 SDK, with the arm in position control and the base in velocity control. Although the Ranger Mini 3 supports Ackermann steering, omnidirectional translation, and in-place rotation, we use only translation mode to avoid control latency from mode switching. 



\subsection{Evaluation of MobRT Benchmark}
In this section, we focus on two main objectives: (1) Evaluating the effectiveness of our data generation framework specifically designed for mobile robots, and assesses whether the generated simulation data can effectively support task execution; (2) Comparing our proposed algorithm with existing baseline methods to demonstrate its superior performance in mobile manipulator tasks.

\subsubsection{Baseline Performance}
We first evaluated four baseline algorithms — ACT, DP, DP3, and iDP3 — across three representative tasks in our benchmark. For each task, policies were trained with varying amounts of expert data: 50, 100, and 200 demonstration sets. To evaluate performance, we executed each trained policy for 30 rollouts under each training condition and recorded the corresponding success rate, thereby constructing a comprehensive benchmark leaderboard.

\begin{table}[h]
    \centering
    \caption{Success rates (\%) of baseline algorithms across different tasks and expert demonstration amounts.}
    \label{tab:benchmark_baseline_results}
    \resizebox{\linewidth}{!}{
        \begin{tabular}{lccccc}
        \toprule
        \multirow{2}{*}{Task Name} & \multirow{2}{*}{Demos} & \multicolumn{4}{c}{Algorithm} \\
        \cmidrule(lr){3-6}
        & & ACT & DP & DP3 & iDP3 \\
        \midrule
        \multirow{3}{*}{Open Cabinet Drawer} 
        & 50  & \percentage{7}{30}           & \textbf{\percentage{8}{30}}  & \textbf{\percentage{8}{30}}  & \percentage{5}{30}  \\
        & 100 & \textbf{\percentage{15}{30}} & \percentage{11}{30}          & \percentage{8}{30}           & \percentage{8}{30}  \\
        & 200 & \textbf{\percentage{18}{30}} & \percentage{15}{30}          & \percentage{16}{30}          & \percentage{13}{30} \\
        \midrule
        \multirow{3}{*}{Container Place} 
        & 50  & \percentage{0}{30}           & \textbf{\percentage{1}{30}} & \textbf{\percentage{1}{30}} & \percentage{0}{30} \\
        & 100 & \textbf{\percentage{9}{30}}  & \percentage{6}{30}          & \percentage{0}{30}          & \percentage{1}{30} \\
        & 200 & \textbf{\percentage{21}{30}} & \percentage{14}{30}         & \percentage{4}{30}          & \percentage{1}{30} \\
        \midrule
        \multirow{3}{*}{Open Dish Washer} 
        & 50  & \textbf{\percentage{9}{30}}  & \percentage{6}{30}  & \textbf{\percentage{9}{30}}  & \percentage{3}{30} \\
        & 100 & \textbf{\percentage{15}{30}} & \percentage{9}{30}  & \percentage{8}{30}           & \percentage{6}{30} \\
        & 200 & \textbf{\percentage{15}{30}} & \percentage{12}{30} & \percentage{10}{30}          & \percentage{7}{30} \\
        \midrule
        \multirow{3}{*}{\textbf{\textit{Average}}} 
        & 50  & \FPeval{\avgRGB}{round((7+0+9)/3/30*100,1)}\avgRGB\% & \FPeval{\avgPC}{round((8+1+6)/3/30*100,1)}\avgPC\% & \FPeval{\avgRGB}{round((8+1+9)/3/30*100,1)}\textbf{\avgRGB\%} & \FPeval{\avgPC}{round((5+0+3)/3/30*100,1)}\avgPC\% \\
        & 100 & \FPeval{\avgRGB}{round((15+9+15)/3/30*100,1)}\textbf{\avgRGB\%} & \FPeval{\avgPC}{round((11+6+9)/3/30*100,1)}\avgPC\% & \FPeval{\avgRGB}{round((8+0+8)/3/30*100,1)}\avgRGB\% & \FPeval{\avgPC}{round((8+1+6)/3/30*100,1)}\avgPC\% \\
        & 200 & \FPeval{\avgRGB}{round((18+21+15)/3/30*100,1)}\textbf{\avgRGB\%} & \FPeval{\avgPC}{round((15+14+12)/3/30*100,1)}\avgPC\% & \FPeval{\avgRGB}{round((16+4+10)/3/30*100,1)}\avgRGB\% & \FPeval{\avgPC}{round((13+1+7)/3/30*100,1)}\avgPC\% \\
        \bottomrule 
    \end{tabular}
    }
\end{table}

\begin{figure}[!htbp]
    \centering
    \includegraphics[width=0.475\textwidth]{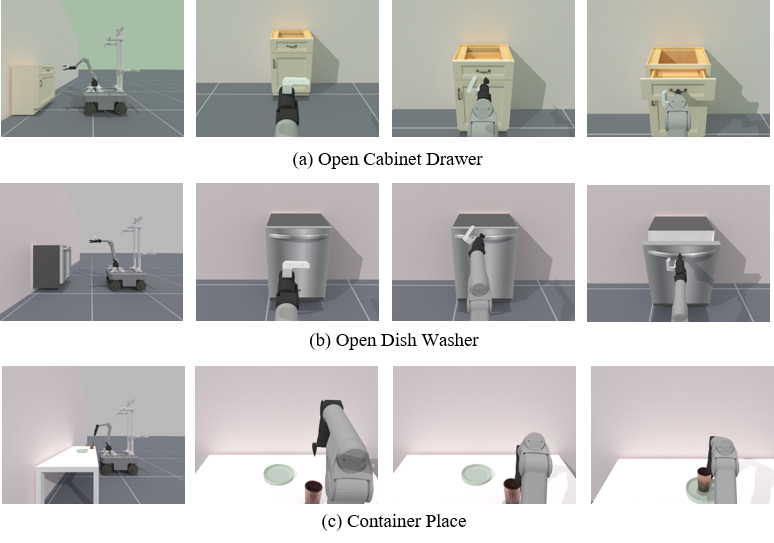}
    \caption{\textbf{Task Execution of MobRT in Simulation.} Representative tasks, including drawer and dishwasher opening and object placement, demonstrate MobRT’s whole-body coordination and sequential manipulation.}
    \label{fig:task_exec}
    \vspace{-0.2cm}
\end{figure}

The experimental results, summarized in Table~\ref{tab:benchmark_baseline_results}, elucidate the comparative performance of each baseline algorithm. As anticipated, an increase in the number of expert demonstrations consistently yielded enhanced performance across all evaluated tasks. For example, in the “Open Cabinet Drawer” task, the ACT algorithm's success rate improved from 0\% with 50 demonstrations to 70\% with 200 demonstrations. Comparable trends were observed across other baseline methods and tasks, demonstrating a strong positive correlation between the volume of expert demonstrations and task success rates, except for DP3 and iDP3 on the \textit{Container Place} task; the underlying reasons for this exception will be discussed in \cref{tab:pointcloud_foreground}. These findings not only validate the effectiveness of expert data automatically generated by MobRT in enhancing performance on mobile manipulation platforms but also highlight the fundamental role of sufficient training data in the development of robust policies for complex robotic tasks.

\subsubsection{Performance of Our Algorithm}

\begin{table}[h]
    \centering
    \caption{Success rates (\%) of ours algorithms across different tasks and expert demonstration amounts.}
    \label{tab:benchmark_ours_results}
    \begin{tabular}{lccc}
        \toprule
        Task Name & Demos & RGB-based & Point Cloud-based \\
        \midrule
        \multirow{3}{*}{Open Cabinet Drawer} 
        & 50  & \textbf{\percentage{14}{30}}  & \percentage{10}{30} \\
        & 100 & \textbf{\percentage{16}{30}}  & \percentage{12}{30} \\
        & 200 & \textbf{\percentage{19}{30}}  & \percentage{15}{30} \\
        \midrule
        \multirow{3}{*}{Container Place} 
        & 50  & \percentage{10}{30}           & \textbf{\percentage{15}{30}} \\
        & 100 & \percentage{22}{30}           & \textbf{\percentage{23}{30}} \\
        & 200 & \textbf{\percentage{26}{30}}  & \percentage{23}{30} \\
        \midrule
        \multirow{3}{*}{Open Dish Washer} 
        & 50  & \textbf{\percentage{18}{30}} & \percentage{12}{30} \\
        & 100 & \textbf{\percentage{22}{30}} & \percentage{10}{30} \\
        & 200 & \textbf{\percentage{24}{30}} & \percentage{17}{30} \\
        \midrule
        \multirow{3}{*}{\textbf{\textit{Average}}} 
        & 50  & \FPeval{\avgRGB}{round((14+10+18)/3/30*100,1)}\textbf{\avgRGB\%}\improvement{26.7} & \FPeval{\avgPC}{round((10+15+12)/3/30*100,1)}\avgPC\%\improvement{21.1} \\
        & 100 & \FPeval{\avgRGB}{round((16+22+22)/3/30*100,1)}\textbf{\avgRGB\%}\improvement{23.4} & \FPeval{\avgPC}{round((12+23+10)/3/30*100,1)}\avgPC\%\improvement{6.7} \\
        & 200 & \FPeval{\avgRGB}{round((19+26+24)/3/30*100,1)}\textbf{\avgRGB\%}\improvement{16.7} & \FPeval{\avgPC}{round((15+23+17)/3/30*100,1)}\avgPC\%\improvement{1.1} \\
        \bottomrule
    \end{tabular}
\end{table}

We then evaluated our proposed algorithms under the same experimental settings. Results are reported in \cref{tab:benchmark_ours_results}. Compared with the baseline methods in \cref{tab:benchmark_baseline_results}, our approach demonstrates consistently higher success rates across all tasks and data regimes. For instance, 
in the \textit{Open Dish Washer} task with 50 demonstrations, the best-performing baseline achieves 30\% success, whereas our method reaches 60\%, yielding a 30-point improvement. Similarly, in the \textit{Container Place} task with 100 demonstrations, our point cloud-based variant improves success rates from 30\% to 76.7\%, representing a 46.7-point gain. On average, both the RGB- and point cloud-based variants exhibit clear advantages over the baselines across different demonstration budgets. The performance gains are especially notable in low-data regimes, which highlights the robustness and data efficiency of our approach.

Overall, these experiments confirm that MobRT-generated data is sufficiently diverse for robust policy learning and that our algorithms consistently outperform existing baselines in complex mobile manipulation tasks.

\subsubsection{Impact of Point Cloud Preprocessing}
In our evaluation, we found that certain baseline algorithms, particularly DP3 and iDP3, showed limited improvement in the \textit{Container Place} task even with more demonstration data. As shown in \cref{tab:pointcloud_foreground}, these point cloud-based methods are highly sensitive to background noise and clutter, which confines their effectiveness mainly to tabletop scenarios and limits their applicability on mobile robotic platforms. In contrast, our proposed point cloud-based policy tokenizes point cloud data rather than relying on max pooling, as done in DP3 and iDP3, thereby preserving spatial details and enabling robust performance across diverse and unstructured environments.

\begin{table}[h]
    \centering
    \caption{Success rates (\%) of DP3 and iDP3 with and without foreground segmentation in the \textit{Container Place} task.}
    \label{tab:pointcloud_foreground}
    \begin{tabular}{lcccc}
        \toprule
        Demos & DP3 & DP3$^*$ & iDP3 & iDP3$^*$ \\
        \midrule
        50  & \percentage{1}{30}  & \textbf{\percentage{13}{30}}\improvement{40} & \percentage{0}{30} & \textbf{\percentage{7}{30}}\improvement{23.3}  \\
        100 & \percentage{0}{30}  & \textbf{\percentage{18}{30}}\improvement{60} & \percentage{1}{30} & \textbf{\percentage{9}{30}}\improvement{26.7}  \\
        200 & \percentage{4}{30}  & \textbf{\percentage{22}{30}}\improvement{60} & \percentage{1}{30} & \textbf{\percentage{18}{30}}\improvement{56.7} \\
        \bottomrule
        \multicolumn{5}{l}{\footnotesize $^*$ with foreground segmentation} \\
    \end{tabular}
\end{table}

\begin{figure}[!htbp]
    \centering
    \includegraphics[width=0.5\textwidth]{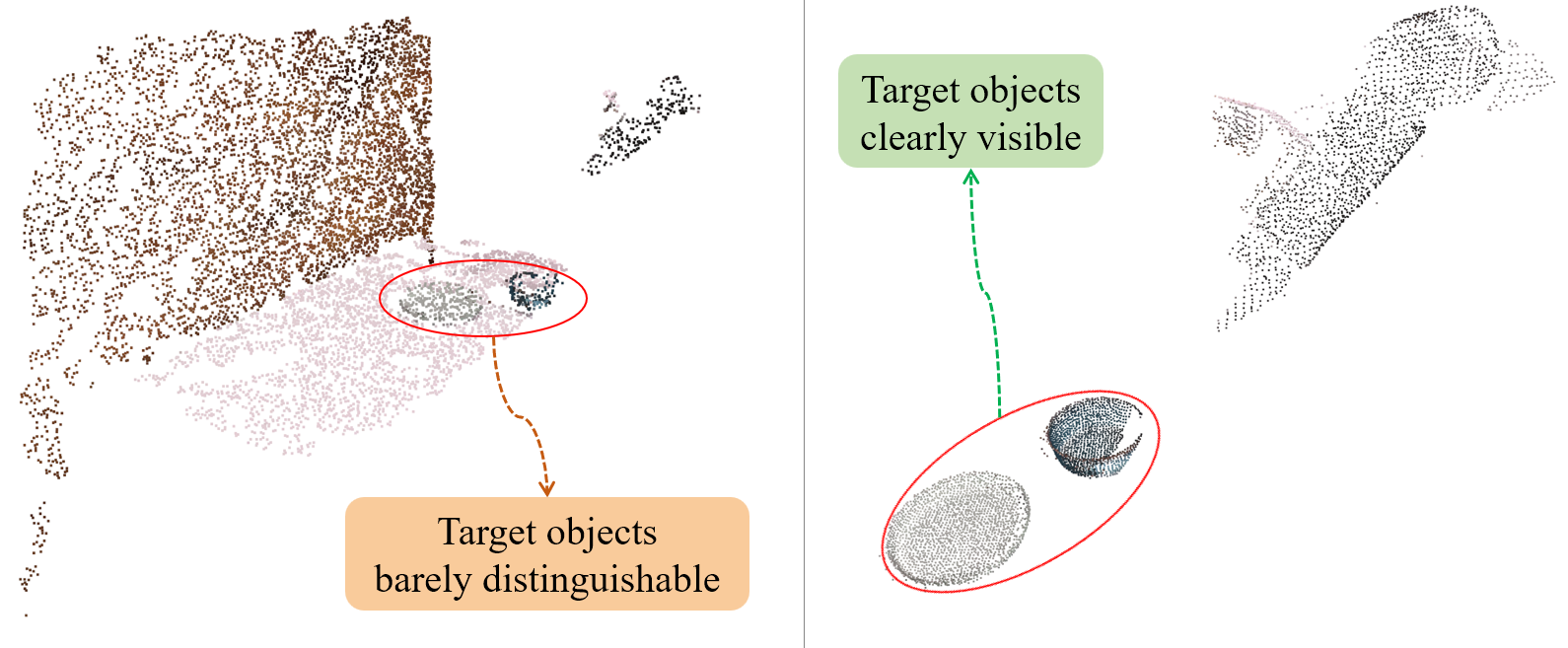}
    \caption{\textbf{Effect of Point Cloud Preprocessing.} Preprocessing enhances object visibility in point clouds, improving downstream policy learning.} 
    \label{fig:pointcloud}
\end{figure}

\subsection{Evaluation of Mixed Data Co-Training}
In this section, we evaluate the effectiveness of MobRT in improving policy robustness in real-world scenarios. To this end, we conduct controlled experiments on the \textit{Open Cabinet Drawer} task. The experiments are performed on our real-world mobile manipulator platform using iDP3 and our algorithm as policy backbones. For each setup, we compare two training conditions: (1) policies trained on 20 real-world demonstrations, and (2) the same real-world data augmented with 300 synthetic trajectories generated by MobRT. Each trained policy is evaluated over 10 rollouts, and the averaged success rates are reported in Table~\ref{tab:hybrid_train_results}.

\begin{table}[h]
    \centering
    \caption{Success rates (\%) on real-world task when trained with different data compositions.}
    \label{tab:hybrid_train_results}
    \resizebox{\linewidth}{!}{
        \begin{tabular}{llcc}
            \toprule
            Task Name & Training Set & iDP3 & Point  Cloud-based \\
            \midrule
            \multirow{2}{*}{Open Cabinet Drawer}
            & Real only         & \percentage{1}{10} & \textbf{\percentage{2}{10}} \\
            & Real + Simulated & \percentage{4}{10} & \textbf{\percentage{6}{10}} \\
            \bottomrule
        \end{tabular}
    }
\end{table}

\begin{figure}[!htbp]
    \centering
    \includegraphics[width=0.48\textwidth]{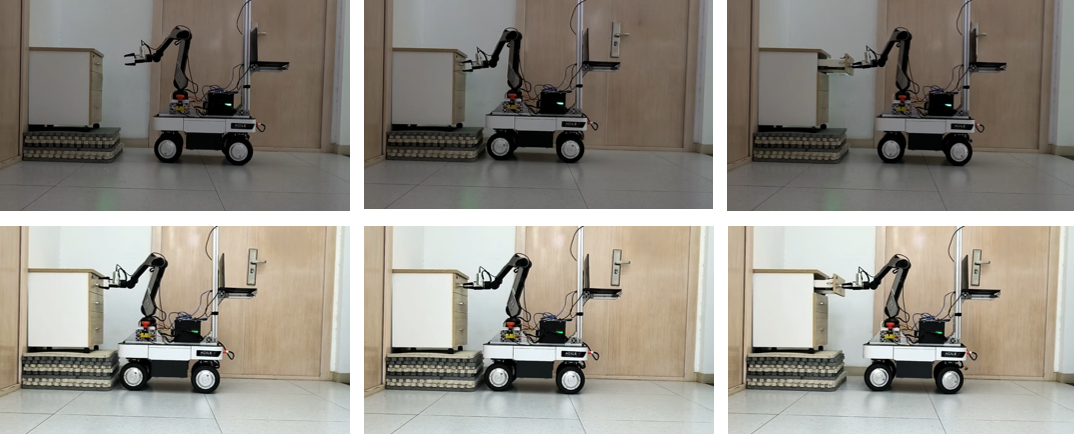}
    \caption{\textbf{Task Execution in Real-World Experiment.} Trained policies enable successful cabinet drawer opening under varying conditions.} 
    \label{fig:task_exec_real}
\end{figure}

Our experimental results show that combining a small number of real demonstrations with MobRT-generated trajectories leads to performance improvements. These improvements can be attributed to MobRT’s modeling of real-world sensing conditions together with extensive domain randomization during data generation, which jointly enhance policy robustness and mitigate overfitting to simulation-specific artifacts. Moreover, we observe that the trained policies exhibit a certain degree of adaptability to variations in environmental lighting as well as to changes in cabinet drawer height, as illustrated in \cref{fig:task_exec_real}.

\section{Conclusions And Limitations}

In summary, we present \textbf{MobRT}, a benchmark and data generation framework for mobile manipulation. MobRT enhances sim-to-real transfer through sensor noise modeling and domain randomization. Our multi-modal policy outperforms baselines in both performance and data efficiency. It should be noted that the current set of tasks covered by MobRT remains limited, and real-world performance has yet to be fully explored. Future work will expand the range and horizon of tasks and integrate reinforcement learning to further improve robustness and adaptability in real-world scenarios.









\bibliographystyle{plain}
\bibliography{references}
\end{document}